\documentclass[lettersize,journal]{IEEEtran}
\usepackage{amsmath,amsfonts}
\usepackage{algorithmic}
\usepackage{algorithm}
\usepackage{array}
\usepackage[caption=false,font=normalsize,labelfont=sf,textfont=sf]{subfig}
\usepackage{textcomp}
\usepackage{stfloats}
\usepackage{url}
\usepackage{verbatim}
\usepackage{graphicx}
\usepackage{cite}
\usepackage{balance}
\usepackage{booktabs}
\usepackage{array}
\usepackage[table,xcdraw]{xcolor}

\usepackage{hyperref}
\hypersetup{colorlinks=true, linkcolor=red, citecolor=blue, urlcolor=magenta}

\begin{document}

\title{DeProPose: Deficiency-Proof 3D Human Pose Estimation via Adaptive Multi-View Fusion}

\author{Jianbin Jiao$^{1}$, Xina Cheng$^{1,*}$, Kailun Yang$^{2}$, Xiangrong Zhang$^{1}$, and Licheng Jiao$^{1}$
        % <-this % stops a space
%
\thanks{This work was supported in part by the National Natural Science Foundation of China (NSFC) under Grant No. 62473139.}%
\thanks{$^{1}$The authors are with the School of Artificial Intelligence, Xidian University, China.}
\thanks{$^{2}$The author is with the School of Robotics, Hunan University, China.}
\thanks{$^{*}$Corresponding author: Xina Cheng.}%
}

% The paper headers
\markboth{}%
{Jiao \MakeLowercase{\textit{et al.}}: DeProPose}

\maketitle

\begin{abstract}
3D human pose estimation has wide applications in fields such as intelligent surveillance, motion capture, and virtual reality. However, in real-world scenarios, issues such as occlusion, noise interference, and missing viewpoints can severely affect pose estimation. To address these challenges, we introduce the task of Deficiency-Aware 3D Pose Estimation. Traditional 3D pose estimation methods often rely on multi-stage networks and modular combinations, which can lead to cumulative errors and increased training complexity, making them unable to effectively address deficiency-aware estimation. To this end, we propose DeProPose, a flexible method that simplifies the network architecture to reduce training complexity and avoid information loss in multi-stage designs. Additionally, the model innovatively introduces a multi-view feature fusion mechanism based on relative projection error, which effectively utilizes information from multiple viewpoints and dynamically assigns weights, enabling efficient integration and enhanced robustness to overcome deficiency-aware 3D Pose Estimation challenges. Furthermore, to thoroughly evaluate this end-to-end multi-view 3D human pose estimation model and to advance research on occlusion-related challenges, we have developed a novel 3D human pose estimation dataset, termed the Deficiency-Aware 3D Pose Estimation (DA-3DPE) dataset. This dataset encompasses a wide range of deficiency scenarios, including noise interference, missing viewpoints, and occlusion challenges, thereby establishing a comprehensive benchmark for evaluating model robustness and performance under various demanding conditions. Experimental results demonstrate that DeProPose exhibits exceptional robustness and accuracy in deficiency-aware scenarios.  Compared to state-of-the-art methods, DeProPose not only excels in addressing the deficiency-aware problem but also shows improvement in conventional scenarios, providing a powerful and user-friendly solution for 3D human pose estimation. The source code will be available at \url{https://github.com/WUJINHUAN/DeProPose}.
\end{abstract}


\section{Introduction}

\IEEEPARstart{H}{uman} pose estimation, due to its high accuracy, strong environmental adaptability, and excellent anti-counterfeiting capabilities, holds significant application value in fields such as security, healthcare, entertainment, sports, industry, and smart living~\cite{10584563,7466118,9350298,9829868,9153745}.
However, the realization of this technology often depends on large and complex datasets, which introduces many challenging issues, such as data scarcity, annotation difficulty, non-rigid motion, and individual differences. 
As a result, many effective methods have been developed to tackle these challenging problems in 3D human pose estimation~\cite{moreno20173d,srivastav2024selfpose3d,zhang2020learning}.
In the task of 3D human pose estimation, the quality of generated image data may fluctuate due to variations in camera hardware performance and environmental factors, making it susceptible to interference from sensor noise, viewpoint changes, and image distortion. 
These interferences can lead to the occurrence of occlusion, data missing, and noise, which represent typical examples of deficiency-aware scenarios.
Such issues can reduce the overall accuracy of 3D pose estimation and present significant challenges in real-world application scenarios. Therefore, it is essential and crucial to develop models or systems that can effectively reduce noise and occlusion interference, while enhancing performance under these conditions.

\begin{figure}[!t] % 使用 figure* 跨两栏
    \centering
    \includegraphics[width=\linewidth]{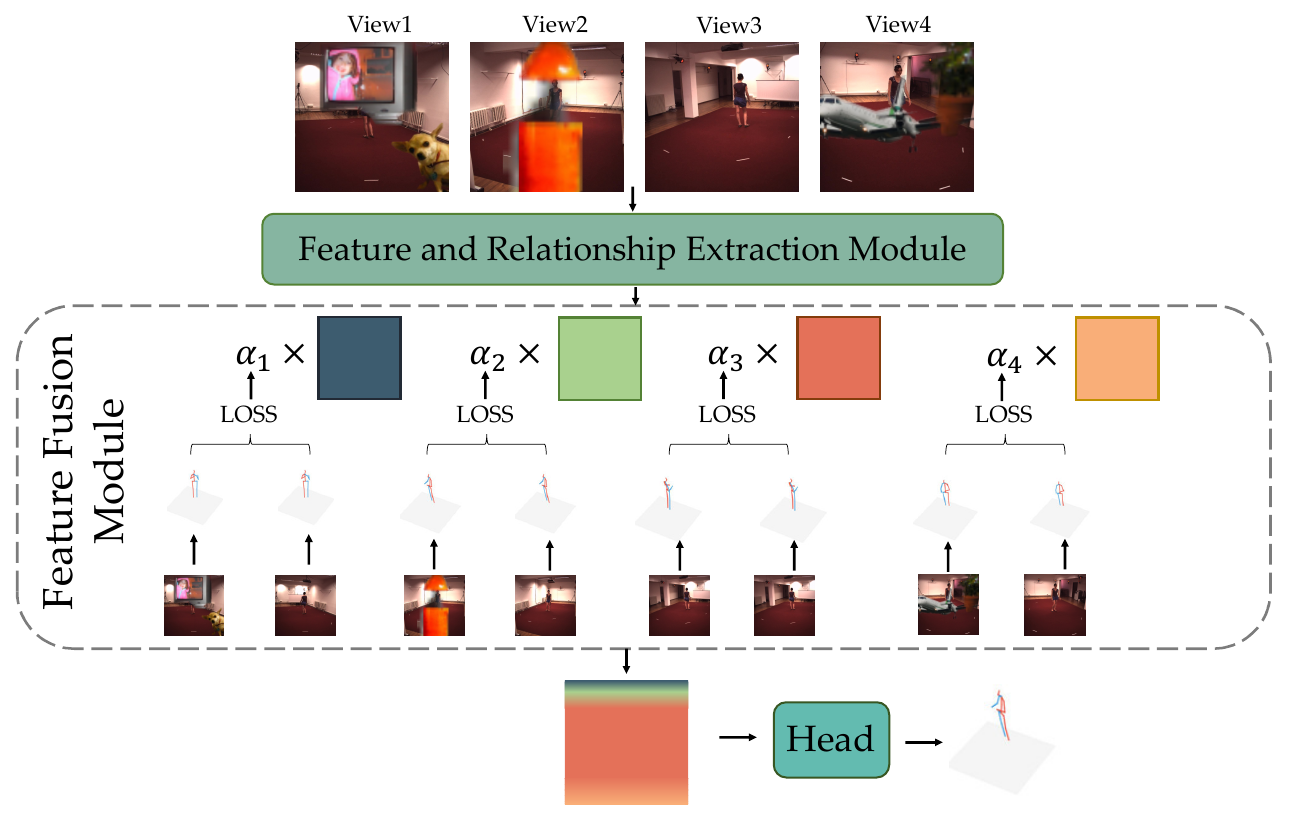} % 设置图片宽度
    \caption{Illustration of the proposed framework for multi-view 3D human pose estimation. 
    Images from multiple views (View1, View2, View3, View4) are processed by the feature and relationship extraction module to extract features. 
    These features are fused in the feature fusion module with adaptive weights ($\alpha_1$, $\alpha_2$, $\alpha_3$, $\alpha_4$) to compute losses. 
    The fused features are then passed to the head to predict the final 3D pose. 
    This design leverages multi-view information and adaptively balances contributions from different views.}
    \label{概念图}
\end{figure}

Although several 3D human pose estimation models have been proposed, most of them primarily focus on single-view scenarios~\cite{liu2020attention,chen2021anatomy,wang2020motion,zeng2020srnet,zheng2021_3d_hpe}, failing to fully exploit the feature information from multi-view datasets. 
This limitation not only hinders improvements in prediction performance but also leads to a significant drop in model performance when faced with deficiency-aware scenarios.
To address this issue, researchers have shifted their focus to multi-view 3D human pose estimation~\cite{liu2021dual,bartol2022generalizable,iskakov2019learnable,zhao2023triangulation,jiang2023probabilistic,jenni2020self,wang2022smart,reddy2021tessetrack,wan2023view,xia2022vitpose}, aiming to enhance pose estimation accuracy by fusing information from different viewpoints. 
However, how to effectively integrate these multi-view features and fully leverage the complementary nature of different viewpoints remains a core challenge in multi-view 3D pose estimation.

Traditional multi-view methods~\cite{kadkhodamohammadi2017multi} often rely on simple stacking or averaging strategies to combine features from different viewpoints. 
Although these approaches are easy to implement, they fail to fully exploit the complementary information between viewpoints and may even introduce redundant information, leading to decreased model accuracy. 
Since these methods do not effectively distinguish the importance of features from different views, simple fusion techniques may transmit redundant or noisy information into the model, negatively affecting the quality and expressiveness of the features. 
This is particularly problematic in complex environments where the complementary nature between viewpoints is not adequately utilized, thus limiting the model's performance.

\begin{figure}[!t] 
    \centering
    \vspace{0.5cm}
    \includegraphics[width=\linewidth]{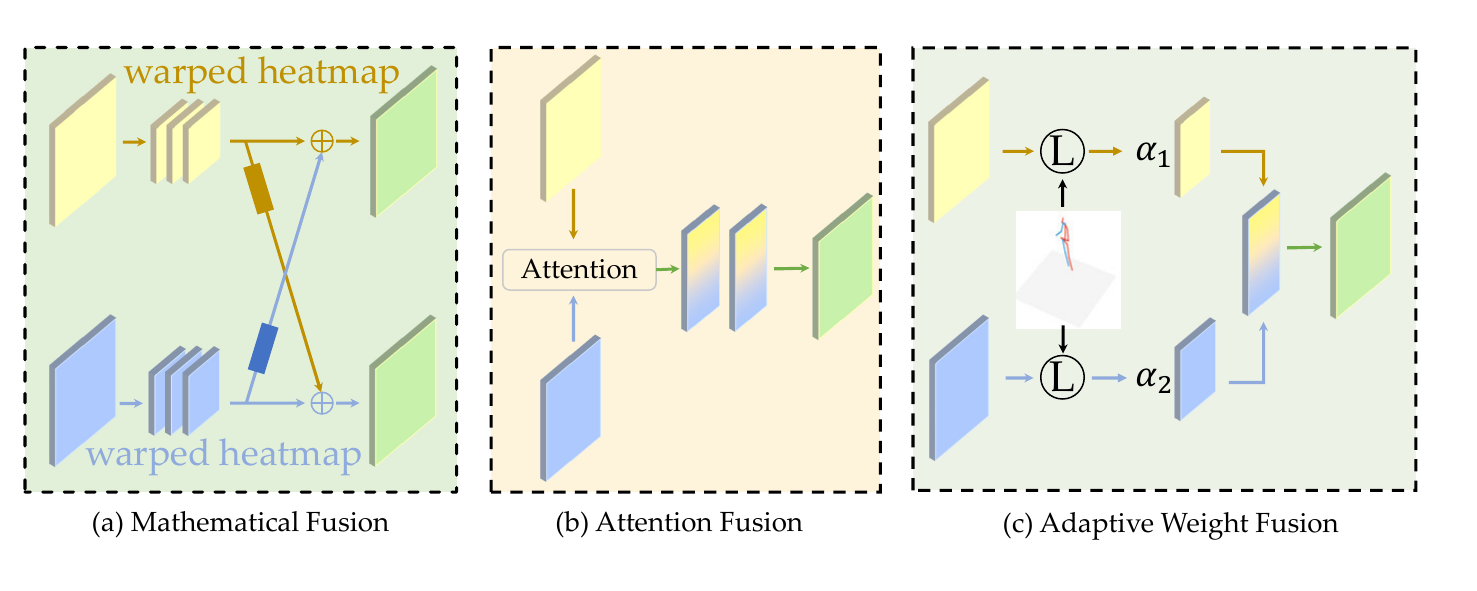} % 设置图片宽度
    \caption{Comparison of different fusion methods. 
    (a) Mathematical fusion~\cite{amin2013multi,solichah2020marker,kadkhodamohammadi2021generalizable,wan2023view,bartol2022generalizable,jiang2023probabilistic}: Data is processed based on the principles of epipolar geometry, and fusion is achieved by analyzing the geometric relationships between data from different perspectives.
    (b) Attention fusion~\cite{cai2024fusionformer,shuai2022adaptive}: Important data is focused on and selected for fusion using the attention mechanism.
    (c) Our adaptive weight fusion: Data fusion is achieved by adaptively adjusting the weights of different data sources. 
    It can handle data flexibly according to data characteristics and make full use of the advantages of each data source.}
    \label{融合模块}
\end{figure}

In recent years, as shown in Fig.~\ref{融合模块}, despite notable progress in multi-view fusion~\cite{bartol2022generalizable,jiang2023probabilistic}, most approaches still focus on ideal scenarios without any occlusion and noise, neglecting the impact of view-incomplete and degraded conditions on performance, which limits their effectiveness in complex environments.
Moreover, most 3D human pose estimation models adopt a two-stage approach~\cite{martinez2017simple}: first performing 2D pose estimation and then mapping it to 3D space. 
While effective in certain scenarios, it relies on multi-stage modular designs, combining CNN~\cite{xu2024multi,haque2016towards}, LSTM~\cite{nunez2019multiview,cheng20203d}, GCN~\cite{jia20233d,liu2021graph,yan2018spatial,wu2021graph}, and other modules, leading to high computational burdens and long processing times, making it difficult to meet real-time requirements. 
The high complexity of the system demands extensive hyperparameter tuning, and the strong dependencies between modules and the problem of information loss can also affect overall performance. 
Furthermore, inconsistent optimization objectives between modules and issues related to information redundancy further limit accuracy. 
Therefore, developing a simple and efficient multi-view 3D pose estimation model that can effectively address deficiency-aware estimation has become an important challenge.

To address the aforementioned challenges, we propose an end-to-end 3D human pose estimation model called the Deficiency-Proof 3D Pose Estimation Model (DeProPose). 
DeProPose aims to build an efficient framework that can effectively handle various types of deficiency-aware problems.
Unlike two-stage methods, DeProPose directly extracts 3D pose features from images, simplifying the model architecture, reducing redundant information transmission, and lowering the cost of hyperparameter tuning and training. 
By utilizing efficient feature extraction and self-attention mechanisms, the model can capture both spatial and temporal relationships in multi-view images.
To tackle the information utilization and deficiency-aware issues in multi-view fusion, we introduce a feature fusion mechanism based on projection error and absolute error.
As shown in Fig.~\ref{概念图}, this mechanism adaptively adjusts feature weights based on the error distribution across different viewpoints, allowing for precise extraction of pose information. 
Our method not only reduces redundant information but also improves estimation accuracy in complex scenarios.
Finally, through multi-view feature fusion, DeProPose effectively handles the deficiency-aware estimation by automatically focusing on viewpoints with less interference, ensuring high-accuracy pose recovery even in complex environments.

Additionally, due to the limited exploration of the deficiency-aware problem in the multi-view human pose domain, we have generated a new dataset Deficiency-Aware 3D Pose Estimation Dataset (DA-3DPE Dataset) specifically designed to address the challenges in multi-view 3D human pose estimation.
This dataset covers three key issues: missing data, noise interference, and viewpoint occlusion. 
These problems frequently occur in complex real-world scenarios, severely impacting the accuracy and robustness of existing methods.
Unlike existing datasets, the DA-3DPE dataset focuses on providing more realistic and challenging samples for multi-view 3D pose estimation, particularly in cases where incomplete viewpoints or data inconsistencies arise during human pose recognition.
To date, no dataset comprehensively addresses these challenges, making the release of this dataset of significant research value and application potential.
The main contributions of this paper can be summarized as follows:
\begin{itemize}
    \item We propose a simple and efficient end-to-end 3D human pose estimation model capable of effectively handling multi-view 3D pose estimation tasks in occlusion and noisy environments, demonstrating strong generalizability and adaptability.

    \item We introduce a feature fusion mechanism based on projection error and absolute error that adaptively adjusts the weights of features from different views to precisely extract human pose information, thereby improving 3D pose estimation accuracy in complex scenarios.

    \item We apply the multi-view fusion mechanism to complex scenes, addressing occlusion and noise issues by supplementing lost information from other views and automatically adjusting view weights, thereby maintaining high-precision pose estimation. 

    \item We introduce the Deficiency-Aware 3D Pose Estimation (DA-3DPE) dataset, a fresh dataset designed to address the challenges of missing data, noise interference, and view occlusion in multi-view 3D human pose estimation.
\end{itemize}

\section{Related Work}
\subsection{Multi-view 3D Human Pose Estimation Model}
Most multi-view 3D human pose estimation methods adopt a two-stage processing framework. 
These methods first estimate 2D human poses from images of each view, then map the 2D poses from multiple views into 3D space to construct the overall 3D pose structure. 
The advantage of this approach lies in leveraging existing 2D pose estimation techniques and achieving high initial 2D detection accuracy. 
Shuai~\textit{et al.}~\cite{shuai2022adaptive} introduce the MTF-Transformer, which employs feature extractors, multi-view fusion transformers, and temporal fusion transformers to adaptively handle varying views and video lengths. 
This approach excels in processing uncalibrated multi-view sequences, effectively improving the model's adaptability and accuracy across diverse scenarios.
Similarly, Cai~\textit{et al.}~\cite{cai2024fusionformer} propose FusionFormer, a concise unified feature fusion transformer. 
By leveraging a unified feature fusion scheme to integrate multi-frame and multi-view features, FusionFormer not only reduces the impact of depth uncertainty but also achieves efficient 3D pose estimation with a compact model size and low computational cost, offering new perspectives for model optimization and practical applications. 
Zhao~\textit{et al.}~\cite{zhao2023triangulation} a triangulation residual loss is proposed for multi-view 3D pose estimation. 
By utilizing multi-view geometric consistency for self-supervised training, this approach addresses the challenge of limited annotated data, providing an effective solution to data scarcity.

With the development of deep learning, end-to-end multi-view 3D pose estimation models have become a research hotspot. 
Unlike traditional methods that rely on intermediate 2D pose estimations, end-to-end methods use a unified network to directly estimate 3D poses from multi-view images. 
This approach allows the model to automatically learn the relationships between features from different views, directly optimizing the 3D pose results, thus reducing information loss and error accumulation during data transmission. 
Pavlakos~\textit{et al.}~\cite{pavlakos2017coarse} propose a method for 3D human pose estimation from a single RGB image, formulating it as a keypoint localization problem in discrete space.
A coarse-to-fine prediction scheme is employed to address high-dimensional challenges, achieving superior performance over existing methods.
The MvP model~\cite{zhang2021direct} is introduced for multi-view, multi-person 3D pose estimation. 
This model directly regresses the 3D poses of multiple individuals from multi-view images using a carefully designed query embedding scheme and projection attention mechanism to efficiently fuse multi-view information. 
The approach demonstrates strong performance and efficiency on multiple benchmark datasets, providing a more effective solution for multi-view, multi-person 3D pose estimation.

Despite their achievements, these methods fail to address application scenarios involving occlusions and often feature complex model mechanisms, limiting their practical applicability and broader usage.

\subsection{Multi-view Feature Fusion Strategy}
Multi-view feature fusion is a critical component in 3D human pose estimation. Bartol~\textit{et al.}~\cite{bartol2022generalizable} introduce a generalizable random framework for human pose triangulation. In terms of feature fusion, it integrates multi-view information by generating random hypotheses. 
For each joint, a random subset of views is selected, and 3D joint coordinates are obtained through triangulation, forming a 3D human pose hypothesis. A scoring neural network is then used to evaluate these hypotheses. 
The network takes in 3D pose coordinates that are specially normalized, including selecting three specific points for rotation calculation and applying this to the coordinates, as well as extracting body part lengths and concatenating them into vectors. 
The final pose estimate is determined by a weighted averaging strategy, where the hypotheses with higher scores are given greater weight. 
Jiang~\textit{et al.}~\cite{jiang2023probabilistic} focus on uncalibrated multi-view 3D human pose estimation. 
Its probabilistic triangulation module models camera poses using probability distributions, iteratively updating the distribution via Monte Carlo sampling. During inference, the proposed camera pose distributions are maintained, and the network updates the weights by sampling to learn the camera poses from 2D features, achieving multi-view feature fusion. 
In the 3D pose reconstruction network, the results from multiple iterations are weighted and averaged as inputs to obtain the final 3D pose estimate. 

Ma~\textit{et al.}~\cite{ma2023self} design a factorization network to predict normalized 3D human poses and camera viewpoint coefficients. 
It takes two random views of 2D skeletons as input and introduces multi-view information constraints into pose prediction. 
By calculating reprojection errors and consistent factorization losses, the former measures the difference between the input and reprojected 2D poses, and the latter ensures that the normalized 2D poses from different views are consistently reconstructed. 
This enables the network to learn accurate pose representations from multi-view information, thus achieving multi-view feature fusion. 
Xia~\textit{et al.}~\cite{xia2022vitpose} propose a simple feature fusion network that introduces positional information into the Transformer structure through multi-view geometric calibration, allowing the network to perceive the spatial relationships between views. 
During feature fusion, for corresponding feature points from the source and reference views, a line-based fusion method is adopted due to the depth uncertainty of the reference view’s feature points. Additionally, a fusion weight adjustment strategy is employed based on the similarity between the fusion heatmap and the ground truth heatmap, effectively integrating multi-view features and improving pose estimation accuracy.

These methods have made significant progress in multi-view feature fusion, effectively utilizing information from different viewpoints to enhance the accuracy of 3D pose estimation. 
However, they still exhibit some limitations when addressing deficiency-aware problems, including loss of feature information, reduced fusion accuracy between viewpoints, and excessive reliance on assumptions, which lead to inaccurate pose estimation when deficiency-aware issues are severe.

\subsection{Methods for Handling Noise and Occlusion Issues}
Occlusion and noise are major challenges in 3D human pose estimation. 
Zhang~\textit{et al.}~\cite{zhang20233d} introduce the 3D-Aware Neural Body Fitting (3DNBF) framework, which achieves 3D human pose estimation through feature-level synthesis and analysis, providing high robustness to occlusion. 
The framework uses Neural Body Volumes (NBV) as the explicit volumetric base model for the human body, composed of Gaussian ellipsoid kernels. 
This method allows for feature-level rendering of the human body and has several advantages over mesh-based representations. 
Additionally, a contrastive learning framework is used to train the feature extractor, addressing the 2D-3D ambiguity problem.
Yu ~\textit{et al.}~\cite{yu2024yolo} designed the SEAM attention module to enhance the feature learning of occluded faces and introduced the Repulsion Loss to address the face occlusion problem. Additionally, they utilized the information of the effective receptive field to design the anchor for improving the detection accuracy.
Lei Ke ~\textit{et al.}~\cite{ke2021deep} proposed the Bilayer Convolutional Network (BCNet). 
They adopted a bilayer graph convolutional network structure to model the occluder and the occludee respectively and took their interactions into account during the mask regression stage, thus effectively handling the occlusion problem.
Cheng~\textit{et al.}~\cite{cheng2019occlusion} introduce an occlusion-aware deep learning framework that utilizes 2D confidence heatmaps and optical flow consistency constraints to filter out unreliable occlusion estimates of keypoints. 
This framework combines 2D and 3D temporal convolutional networks (TCNs) to handle incomplete 2D keypoints. 
The framework consists of 2D pose estimation (using stacked hourglass networks and optical flow to compute confidence scores, followed by 2D TCN to process incomplete keypoints), 3D pose estimation (using 3D TCNs to obtain 3D poses and calculate multiple losses), and a cylindrical human model (for data augmentation and pose regularization, by projecting keypoint visibility and expanding the dataset). 

These methods provide important insights into improving human pose estimation performance in complex scenarios. 
However, since occlusion issues severely impact 2D detection results, they also lead to a significant drop in 3D detection performance.
Therefore, we bypass the 2D detection results through an end-to-end approach and address the deficiency-aware problem by utilizing adaptive multi-view feature fusion. 
This method enables us to effectively integrate information from different viewpoints, reducing the impact of deficiency-aware issues on the final 3D pose estimation, thereby enhancing the robustness and accuracy of the model in complex scenarios.

\section{Method}
\begin{figure*}[!t] % 使用 figure*
    \centering
    \includegraphics[width=0.9\textwidth]{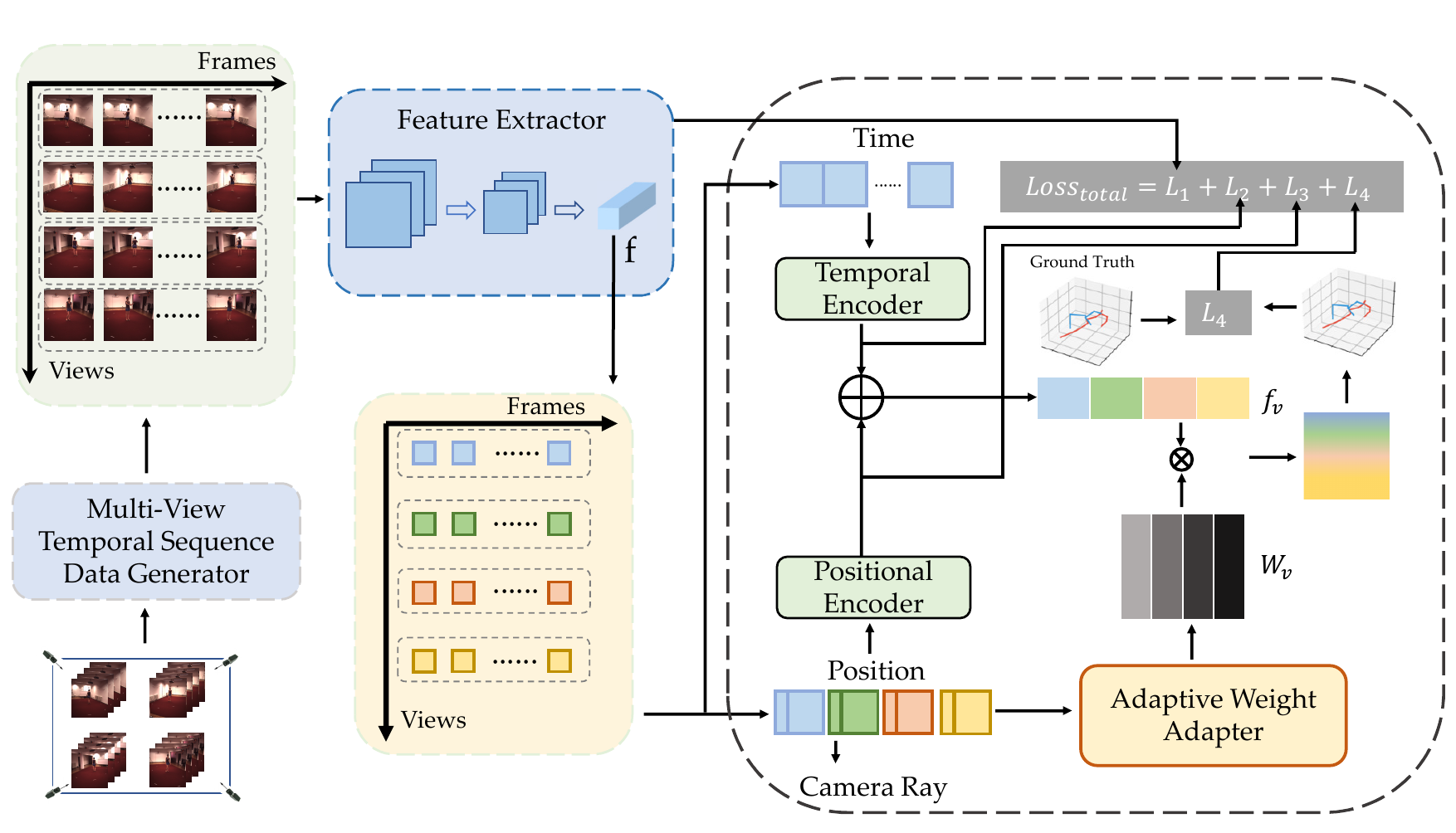} % 设置图片宽度
    \caption{This figure illustrates the architecture of the proposed multi-view temporal 3D human pose estimation model. 
    The pipeline begins with the Multi-View Temporal Sequence Data Generator, which processes input video sequences from multiple views into temporal data frames. 
    These frames are then passed through the Feature Extractor to obtain feature representations ($\mathbf{f}$). 
    The extracted features are fed into the Positional Encoder and Temporal Encoder, which encode spatial and temporal information, respectively. 
    A camera-ray-based positional relationship is incorporated to enhance spatial consistency. 
    An Adaptive Weight Adapter dynamically assigns weights ($\mathbf{w}_v$) to features from different views, enabling adaptive fusion of multi-view information. 
    The fused features ($\mathbf{f}_v$) are used to predict the final 3D pose, which is optimized using a multi-term loss function ($L_{\text{total}} = L_1 + L_2 + L_3 + L_4$). 
    This framework effectively combines spatial-temporal relationships and multi-view adaptive fusion to improve pose estimation accuracy and robustness.}
    \label{结构图}
\end{figure*}

In this section, we provide a detailed introduction and explanation of the proposed method, DeProPose, with the overall framework illustrated in Fig.~\ref{结构图}.

The proposed method consists of three components: the Deficiency-Aware Image Encoder, which leverages a self-attention mechanism~\cite{dosovitskiy2020image} to extract image features, temporal features, and multi-view spatial features, effectively addressing occlusion, noise, and data deficiencies in deficiency-aware scenarios; the Multi-view Feature Fusion Adapter Based on Projection Error and Absolute Error, which dynamically weights and fuses multi-view features to significantly enhance the robustness and accuracy of 3D pose recognition; and the Dataset Generation Strategy and Application of Robust Multi-View Information Fusion, which generates a dataset simulating various defect scenarios and validates the effectiveness and adaptability of the multi-view information fusion strategy in complex environments.

\subsection{Deficiency-Aware Image Encoder}
The motivation for the Deficiency-Aware Image Encoder stems from the need to effectively address occlusion, noise, and data missing issues in Deficient-Aware scenarios. 
This encoder extracts spatial, temporal, and multi-view features from images, providing the foundation for the multi-view feature fusion adapter, thereby enhancing the model's robustness and accuracy in complex scenarios.

In DeProPose, the input video frames come from multiple viewpoints, which are processed by the multi-view time-series data generator. 
The goal of this stage is to convert the images from different viewpoints into a data structure that incorporates both temporal and spatial dimensions, represented as:
\[
\mathbf{X} \in \mathbb{R}^{V \times T \times H \times W \times C},
\]
where \(V\) represents the number of viewpoints, \(T\) denotes the number of time frames, \(H\) and \(W\) are the height and width of each frame, and \(C\) is the number of channels in the image. 
This data structure provides rich spatial and temporal information for subsequent feature extraction and spatial-temporal modeling, enhancing the model's ability to perceive complex scenes. 
Particularly in dynamic environments, it effectively captures the human pose variations from multiple viewpoints.

After data preprocessing, the time-series data is fed into the feature extraction module. 
In this stage, the Swin Transformer model~\cite{liu2021swin} is used to extract high-dimensional features from the time-series data of each viewpoint. 
These features contain critical information about the human pose and can be expressed as:
\[
\mathbf{F} \in \mathbb{R}^{V \times T \times D},
\]
where \(D\) is the dimensionality of the features. This module effectively extracts meaningful pose features from multiple viewpoints and time frames, preserving the detail information to support spatial-temporal modeling and multi-view fusion. The feature extraction process not only enhances the representational capacity of the data but also ensures efficient computation of features, laying a solid foundation for the subsequent stages.

After feature extraction, DeProPose enters the spatial-temporal modeling stage. The main task of this stage is to further enhance the model's expressiveness by capturing the relationships between temporal and spatial information, thus improving the accuracy and robustness of 3D human pose estimation.

The goal of temporal modeling is to capture the dynamics of human pose changes over time. Since human poses are dynamic, traditional single-frame image features cannot accurately capture these temporal variations. Therefore, DeProPose introduces a Temporal Encoder to process time-series data, capturing the dynamic changes of the human body between different time frames. Specifically, the Temporal Encoder processes the time-series data as follows:
\[
\mathbf{F}_t = \text{TemporalEncoder}(\mathbf{F}).
\]
Here, \(\mathbf{F}\) represents the features extracted from the time series of each viewpoint, and \(\mathbf{F}_t\) represents the enhanced features after processing by the Temporal Encoder. 
These enhanced features effectively reflect the dynamic patterns of the human pose as it evolves over time, such as joint movement trajectories and pose transitions. 
This process not only captures the continuity in the temporal dimension but also helps the model understand long-term dependencies, enhancing its ability to adapt to both rapid and slow movements.

In addition to temporal modeling, enhancing spatial information is equally crucial. 
Accurate human pose estimation requires consideration of spatial locations from various viewpoints. Therefore, DeProPose introduces a Positional Encoder to enhance spatial features by incorporating camera ray information. Specifically, the ray information from each viewpoint is represented by the ray angle, which provides additional spatial context for the model, improving its ability to recognize the position of the human body in 3D space:
\[
\mathbf{F}_p = \text{PositionalEncoder}(\mathbf{F}, \mathbf{P}),
\]
where \(\mathbf{P}\) represents the ray information of each viewpoint, and \(\mathbf{F}_p\) denotes the enhanced features with spatial position information. 
The Positional Encoder combines the ray information with the features, enhancing the model's spatial understanding, and enabling it to accurately infer the position of human joints in 3D space. 
This approach helps mitigate the impact of viewpoint changes and effectively addresses potential occlusion problems in multi-view scenarios.

To further improve the model's performance, DeProPose combines both temporal and spatial information for integrated modeling. In real-world scenarios, human poses not only change over time but also have different spatial expressions from different viewpoints. 
The Spatial-Temporal Fusion module integrates the temporally enhanced features (\(\mathbf{F}_t\)) and spatially enhanced features (\(\mathbf{F}_p\)) to provide a more precise and comprehensive pose representation. 
The fusion of temporal and spatial features reflects the dynamic changes of the time series while retaining key spatial information.
The spatial-temporal fusion process is performed as follows:

\[
\mathbf{F}_{tp} = \text{SpatialTemporalFusion}(\mathbf{F}_t, \mathbf{F}_p)
\]

The spatial-temporal fused feature \(\mathbf{F}_{tp}\) combines all the information from both time and space, allowing the model to better adapt to dynamic changes in human poses, especially in fast-paced human motion scenarios, where the effects of spatial-temporal modeling are more pronounced.

\subsection{Multi-view Feature Fusion Adapter Based On Projection Error And Absolute Error}
In the DeProPose framework, multi-view feature fusion serves as a pivotal component for enhancing the model's capability in processing data with diverse deficiency scenarios, including noise interference, missing viewpoints, and occlusion challenges.
By integrating features from different viewpoints, the model can leverage the advantages of each viewpoint, effectively overcoming the limitations posed by a single viewpoint.

\subsubsection{Projection Error Calculation}
For each view \( v \), the predicted 3D pose \( \hat{p}_{3D}^v \) is first projected into the 2D space using the camera projection model, resulting in the projected 2D pose \( \hat{p}_{proj}^v \). 
The projection error is then calculated as:
\[
e_{proj}^v = |\hat{p}_{proj}^v - p_{2D}|,
\]
where \( p_{2D} \) is the ground truth 2D pose (known during training). 
The projection error reflects the level of inaccuracy in the 3D-to-2D projection process. 
A smaller error indicates that the predicted 3D pose in the view is closer to the true 2D pose after projection, and thus, the weight of this view in feature fusion should be higher.

\subsubsection{Absolute Error Calculation}
The absolute error directly measures the difference between the pose features of each view and the ground truth pose features. 
For view \( v \), the absolute error is calculated as:
\[
e_{abs}^v = |f_v - f_{3D}|,
\]
where \( f_v \) represents the pose features extracted from view \( v \), and \( f_{3D} \) represents the ground truth pose features (which can be obtained through data annotations during training). 
The absolute error evaluates the reliability of each view at the feature level. 
It complements the projection error, and together, they determine the weight of each view in the fusion process.

\subsubsection{Weight Calculation and Feature Fusion}
Based on the computed projection error and absolute error, the fusion weight \( \omega_v \) for each view is calculated. 
The weight calculation is as follows:
\[
\omega_v = \frac{1}{e_{proj}^v + e_{abs}^v + \epsilon},
\]
where \( \epsilon \) is a small constant used to prevent division by zero. 
Then, the features \( f_v \) from each view are weighted and fused according to their respective weights to obtain the final fused feature:
\[
F = \sum_{v=1}^V \omega_v f_v,
\]
where \( V \) is the number of views. 
This fusion method adaptively assigns weights based on the quality of each view, effectively integrating multi-view information, reducing the influence of noise and redundant data, and improving the accuracy of pose recognition.

\subsubsection{Error Calculation}
In addition to the projection error and absolute error, the intermediate error is also computed. For the output of each Transformer block, the intermediate error is calculated. 
Let the feature output from the \( k \)-th Transformer block be \( F_k \), then the intermediate error \( E_{mid,k} \) is defined as:
\[
E_{mid,k} = |F_k - F_{true}|.
\]
Finally, the total error is computed by considering the projection error, absolute error, and intermediate error. 
The total error \( E_{total} \) is the sum of these errors:
\[
E_{total} = E_{proj} + E_{abs} + E_{mid}.
\]
By calculating the total error, the performance and optimization direction of the model can be better evaluated.

\subsection{Dataset Generation Strategy and Application of Robust Multi-View Information Fusion}

\subsubsection{Deficiency-Aware 3D Pose Estimation Dataset Generation Strategy}
To better simulate human pose recognition in complex environments, our DA-3DPE dataset generates noisy, missing, and occluded images using three different types of disturbances. 
The noisy images include Gaussian noise, salt-and-pepper noise, and speckle noise. Missing and noise disturbances are randomly applied to one of the four views, while occlusion is randomly applied to images from three of the four views.

Noisy images contain three types of noise: Gaussian noise, salt-and-pepper noise, and speckle noise:

\begin{itemize}
    \item \textbf{Gaussian Noise}: Noise from a standard normal distribution is added to the image:
    \[
    I_{noisy} = I + \mathcal{N}(0, \sigma^2),
    \]
    where \( I \) is the original image, and \( \mathcal{N}(0, \sigma^2) \) is Gaussian noise with mean 0 and variance \( \sigma^2 \).
    
    \item \textbf{Salt-and-Pepper Noise}: Randomly replace pixel values at random locations with black or white:
    \[
    I_{noisy}(x, y) = 
    \begin{cases} 
    0, & \text{with probability } p/2 \\
    255, & \text{with probability } p/2 \\
    I(x, y), & \text{with probability } 1-p
    \end{cases}
    \]
    where \( p \) is the noise density and \( (x, y) \) are the pixel coordinates of the image.

    \item \textbf{Speckle Noise}: Speckle noise is introduced by multiplying the image with noise points:
    \[
    I_{noisy}(x, y) = I(x, y) \cdot (1 + \mathcal{N}(0, \sigma^2)),
    \]
    where \( \mathcal{N}(0, \sigma^2) \) is Gaussian noise, simulating the random variation of speckle noise.
\end{itemize}

Missing images are generated by adding multiple small black blocks at random positions on the image to simulate data loss. 
Specifically, several rectangular regions are randomly chosen, and the pixel values in these regions are set to zero, simulating the loss of partial body information:
\[
I_{missing}(x, y) = 
\begin{cases}
0, & \text{a random black block} \\
I(x, y). & \text{otherwise}
\end{cases}
\]

The positions, sizes, and quantities of the black blocks are determined randomly, simulating the local information loss in the image.

We adopt the method from~\cite{sarandi2018robust} to generate occluded images. 
This approach simulates occlusion by overlaying object images from the Pascal VOC 2012 dataset onto the target images, following semantic segmentation of the objects. 
Specifically, object regions are extracted from the semantic segmentation results, and parts or entire regions of these objects are randomly placed at different positions within the target image to simulate occlusion interference in the environment. 
The degree of occlusion is defined as the percentage of occluded pixels within the human bounding box, and this value is varied between $0\%$ and $70\%$.

\subsubsection{Application Scenarios of Fusion Strategy}

Traditional methods often rely on a single viewpoint for inference, leading to the loss of information about certain joints. 
To address this issue, DeProPose leverages the advantages of multi-view perspectives, using information complementarity between views to recover complete human poses. 
When a joint is occluded in one view, DeProPose can obtain relevant information from other unobstructed views to accurately recover the occluded part. 
For example, if the arm is occluded in the front view, DeProPose can retrieve the pose information of the arm from the side or back view, and integrate this information through the multi-view feature fusion mechanism, ensuring that the 3D pose of the arm is accurately reconstructed.

For deficiency-relevant disturbances, particularly in low-quality images or complex environments, deficiency-aware noise may severely impact pose recognition results. 
To mitigate this issue, DeProPose employs a multi-view feature fusion mechanism that dynamically adjusts the weights to reduce the impact of disturbances. 
For views with higher disturbances, due to larger projection and absolute errors, DeProPose assigns lower weights, thereby minimizing the interference from these views on the final recognition results.

\section{Experiments}

\begin{figure*}[!t] % 使用 figure*
    \centering
    \includegraphics[width=1\textwidth]{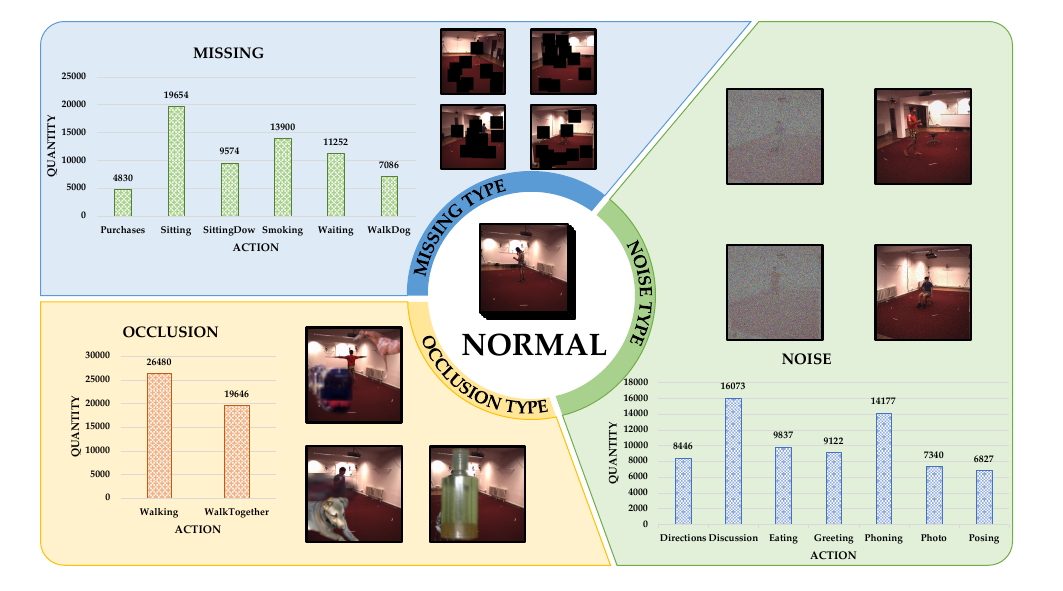} % 设置图片宽度
    \caption{This figure illustrates examples from the DA-3DPE dataset, a multi-view 3D human pose recognition dataset. 
    It includes $401,017$ normal images, $71,823$ noisy images with three types of noise (salt-and-pepper, Gaussian, speckle), $56,724$ missing images with random occlusions, and $46,125$ occluded images where three of four views are obstructed. 
    This dataset provides diverse scenarios to evaluate and improve model robustness under real-world conditions such as noise, missing data, and occlusion.}
    \label{数据集图片}
\end{figure*}

\subsection{Datasets for Evaluation}

To validate the effectiveness of the proposed model, we conducted experiments on several challenging datasets, including the Human3.6M dataset~\cite{ionescu2013human3} and the established Deficiency-Aware 3D Pose Estimation (DA-3DPE) dataset.

\subsubsection{Human3.6M Dataset}
The Human3.6M~\cite{ionescu2013human3} is a large public dataset for 3D human pose estimation research, featuring $3.6$ million images with corresponding 3D human poses. 
The dataset includes $11$ professional actors ($6$ male, $5$ female) and spans $7$ scenes
(such as discussions, smoking, photography, and phone calls). 
Comprising videos captured by $4$ calibrated high-resolution cameras at $50Hz$, the dataset’s labels are derived from precise 3D joint positions and angles obtained from a high-speed motion capture system. 
We utilize five subjects from the dataset for training and reserve two subjects for testing purposes.

\subsubsection{Deficiency-Aware 3D Pose Estimation Dataset}
The established Deficiency-Aware 3D Pose Estimation (DA-3DPE) dataset is a multi-view 3D human pose recognition dataset specifically designed to address common data deficiency challenges in real-world scenarios. 
Based on the widely used Human3.6M~\cite{ionescu2013human3} dataset, it includes four distinct types of images: normal images, occluded images, noisy images, and missing images. 
The dataset introduces challenges such as noise interference, occlusion, and missing data, as illustrated in Fig.~\ref{数据集图片}, which often occur in complex environments and severely affect the accuracy and robustness of existing pose estimation methods.

\textbf{Normal Images:}
These images represent ideal conditions, capturing clear, unobstructed views of human poses from multiple viewpoints. They serve as a baseline for evaluating pose estimation models under standard, noise-free conditions. 
There are $401,017$ normal images in the dataset, all featuring high-quality 3D pose annotations, which are essential for training and testing models in ideal conditions.

\textbf{Occluded Images:}
Occlusion is a significant challenge in human pose recognition, as certain parts of the body may be blocked by objects, other people, or the subject’s own posture. 
The established DA-3DPE dataset simulates occlusion by randomly selecting three out of four viewpoints to be occluded in each image. 
This simulates real-world scenarios where incomplete views of the human body are captured due to environmental factors. 
There are $46,125$ occluded images in the dataset, testing the robustness of pose estimation models when partial information is available.

\textbf{Noisy Images:}
Noise is another common issue in image data, which can degrade the quality of captured data and make pose estimation more challenging. The dataset includes noisy images with three different types of noise:
\begin{itemize}
    \item \textbf{Salt-and-Pepper Noise}: Randomly replacing pixel values with black or white, creating a speckled appearance.
    \item \textbf{Gaussian Noise}: Introducing random variations in pixel intensity based on a Gaussian distribution, causing blur or graininess in the images.
    \item \textbf{Speckle Noise}: Creating grainy patterns in the image, often simulating issues such as sensor malfunctions or poor lighting conditions.
\end{itemize}

In the noisy images, one of the four viewpoints in each image is randomly selected to contain noise. The dataset contains 71,823 noisy images, allowing for testing model performance under various noise conditions.

\textbf{Missing Images:}
Missing data is a common challenge, especially in real-world applications where sensor failures or incomplete coverage of the subject may lead to missing parts of the visual data. 
The DA-3DPE dataset simulates this by randomly selecting one of the four viewpoints in each image to be completely missing. 
This forces the model to infer the missing pose information from the remaining viewpoints. 
There are $56,724$ missing images in the dataset, providing a realistic test scenario for handling incomplete data.

\textbf{Dataset Structure and Usage:}
The DA-3DPE dataset is designed to simulate real-world deficiencies, with each image consisting of four viewpoints where one of the following conditions is applied: normal, occluded, noisy, or missing.
\begin{itemize}
    \item \textbf{Occluded images}: Three out of four viewpoints are randomly occluded.
    \item \textbf{Missing and noisy images}: One of the four viewpoints is randomly selected to be missing or noisy.
\end{itemize}

This dataset provides a comprehensive platform for researchers and developers to train and evaluate models under a variety of challenging conditions, such as occlusion, noise, and missing data. It allows testing model robustness in scenarios involving incomplete or corrupted data, while also enhancing the generalization ability of pose estimation systems by exposing them to realistic data deficiencies. Additionally, the dataset supports the development of algorithms capable of handling incomplete, noisy, or occluded data, which is crucial for practical applications in fields such as surveillance, human-computer interaction, sports analytics, and healthcare.

The DA-3DPE dataset is a valuable resource for advancing research in 3D human pose estimation, providing a realistic set of challenges including missing data, noise interference, and occlusion. By incorporating these common real-world issues, the dataset enables the development of more robust and adaptive pose recognition models. This dataset serves not only as an important contribution to the research community but also has broad potential applications in dynamic and complex environments.

\subsection{Experimental Setup}
Our experiments were conducted using the PyTorch~\cite{paszke2017automatic} framework on an NVIDIA GeForce RTX 3090 GPU. The model was trained using the AdamW~\cite{kingma2014adam} optimizer in conjunction with Vision Transformer~\cite{dosovitskiy2020image} and Swim Transformer~\cite{liu2021swin} architectures. 
The Mean Per Joint Position Error (MPJPE) was used as the loss function, which evaluates the accuracy of predicted joint positions. The model was trained and tested on the Human3.6 dataset, as well as evaluated on our custom occlusion dataset to assess its robustness under challenging conditions. 
All input images were resized and cropped to a resolution of $224{\times}224{\times}3$.

During training, we used a batch size of $32$ and set the total number of training epochs to $100$. The initial learning rate was set to $1e{-}4$, with a learning rate warm-up applied during the first $5$ epochs to stabilize the early training phase. 
To prevent overfitting, weight decay was applied with a value of $0.05$. The minimum learning rate was set to $1e{-}6$. 
For learning rate scheduling, we employed a cosine annealing strategy, where the learning rate was decayed every $10$ epochs according to a pre-defined schedule.

For optimization, we utilized the AdamW optimizer, with Beta parameters set to $0.9$ and $0.999$, and a momentum value of $0.9$. 
Additionally, the model supports automatic recovery during training by saving and loading checkpoints, ensuring that training can resume from interruptions without loss of progress. 
These settings were carefully selected to ensure stable training and optimize the performance of the model.

\subsection{Evaluation Metrics}
In this paper, we use Mean Per Joint Position Error (MPJPE)~\cite{zheng2023deep} and Procrustes Mean Per Joint Position Error (P-MPJPE) as the evaluation metrics to measure the accuracy of the model's predictions in 3D human pose estimation tasks. 

MPJPE and P-MPJPE are commonly used metrics to evaluate the accuracy of 3D human pose estimation models. 
MPJPE measures the overall accuracy of the model by calculating the Euclidean distance between the predicted and ground truth positions for each joint and then averaging across all joints. The formula is as follows:
\[
\text{MPJPE} = \frac{1}{N} \sum_{i=1}^{N} \| \mathbf{P}_i - \mathbf{G}_i \|,
\]
where \(\mathbf{P}_i\) and \(\mathbf{G}_i\) represent the predicted and ground truth positions of the \(i\)-th joint, respectively, and \(N\) is the total number of joints. \(\| \cdot \|\) denotes Euclidean distance, which is used to compute the straight-line distance between two points. For two points in 3D space, \(\mathbf{P}_i = (x_1, y_1, z_1)\) and \(\mathbf{G}_i = (x_2, y_2, z_2)\), the Euclidean distance is computed as:
\[
\| \mathbf{P}_i - \mathbf{G}_i \| = \sqrt{(x_1 - x_2)^2 + (y_1 - y_2)^2 + (z_1 - z_2)^2}.
\]

Although MPJPE is intuitive and easy to compute, it can be affected by factors such as rotation and translation. 
Therefore, P-MPJPE is often used for a more accurate evaluation of the model. 
P-MPJPE extends MPJPE by applying a Procrustes transformation to the predicted and ground truth joint positions, eliminating the effects of rotation and translation, and ensuring that the evaluation reflects only the errors due to pose. The formula is as follows:
\[
\text{P-MPJPE} = \frac{1}{N} \sum_{i=1}^{N} \| \mathbf{P}_i' - \mathbf{G}_i \|,
\]
where \(\mathbf{P}_i'\) represents the predicted joint positions after applying the Procrustes transformation. 
The symbol \(\| \cdot \|\) still denotes the Euclidean distance. 
P-MPJPE provides a more fair and precise measure by removing the impact of pose rotations and translations.

As an evaluation metric, MPJPE and P-MPJPE comprehensively reflect the model's accuracy in global joint positioning, where a smaller value indicates more accurate pose estimation. 
In our experiments, we use MPJPE and P-MPJPE to evaluate our method and compare it with existing approaches to validate the effectiveness of our proposed model.

\subsection{Comparison With Baselines}
In this section, we first evaluate the model on the Human3.6M dataset and compare it with various single-view and multi-view methods. 
The evaluation results are presented using the MPJPE metric, with some results showing the individual performance for each action and others representing the average across all actions. 
The following are the methods used for the comparison:

\begin{table*}[t!]
\scriptsize
\renewcommand{\thetable}{I}
\centering
\caption{The 3D human pose detection results evaluated on the Human3.6M dataset.}
\label{table1}
\resizebox{\linewidth}{!}{
\setlength{\tabcolsep}{3pt}
\begin{tabular}{l||l|ccccccccccccccc|c}
\hline
\rowcolor{gray!20}
\textbf{MPJPE$\downarrow$}       &   \textbf{Venue}        & \textbf{Dir.} & \textbf{Disc.} & \textbf{Eat}  & \textbf{Greet} & \textbf{Phone} & \textbf{Photo} & \textbf{Pose} & \textbf{Purch.} & \textbf{Sit}  & \textbf{SitD.} & \textbf{Somke} & \textbf{Wait} & \textbf{WalkD.} & \textbf{Walk} & \textbf{WalkT.} & \textbf{Average} \\
\hline
\hline
\rowcolor{gray!20}
& \multicolumn{17}{c}{\textbf{Single - View Methods}}\\
\hline
Liu \textit{et al.}~\cite{liu2020attention} &  {CVPR’20}  & 41.8 & 44.8  & 41.1 & 44.9  & 47.4  & 54.1  & 43.4 & 42.2   & 56.2 & 63.6  & 45.3  & 43.5 & 45.3  & 31.3 & 32.2   & 45.1    \\
SRNet~\cite{zeng2020srnet}      &  {ECCV’20}  & 46.6 & 47.1  & 43.9 & 41.6  & 45.8  & 49.6  & 46.5 & 40.0  & 53.4 & 61.1  & 46.1  & 42.6 & 43.1 & 31.5 & 32.6   & 44.8    \\
UGCN~\cite{wang2020motion}       &  {ECCV’20}  & 41.3 & 43.9  & 44.0 & 42.2  & 48.0  & 57.1  & 42.2 & 43.2   & 57.3 & 61.3  & 47.0  & 43.5 & 47.0   & 32.6 & 31.8   & 45.6    \\
Chen \textit{et al.}~\cite{chen2021anatomy}&TCSVT’21 & 42.1 & 43.8 & 41.0 & 43.8  & 46.1  & 53.5  & 42.4 & 43.1   & 53.9 & 60.5 & 45.7  & 42.1 & 46.2   & 32.2 & 33.8   & 44.6    \\
PoseFormer~\cite{zheng2021_3d_hpe} &ICCV’21   & 41.5 & 44.8  & 39.8 & 42.5  & 46.5  & 51.6  & 42.1 & 42.0   & 53.3 & 60.7  & 45.5  & 43.3 & 46.1   & 31.8 & 32.2   & 44.3   \\
\hline
\hline
\rowcolor{gray!20}
& \multicolumn{17}{c}{\textbf{Multi - View Methods}}\\
\hline
Dual-view~\cite{liu2021dual} & IET’21   & 21.4 & 23.1 & 20.7 & 22.1 & 21.5 & 26.5 & 20.5 & 22.3 & 22.8 & 25.9 & 21.2 & 21.1 & 24.9 & 20.9 & 22.2 & 22.5    \\
Bartol \textit{et al.}~\cite{bartol2022generalizable} & CVPR’22   & 27.5 & 28.4 & 29.3 & 27.5 & 30.1 & 28.1 & 27.9 & 30.8 & 32.9 & 32.5 & 30.8 & 29.4 & 28.5 & 30.5 & 30.1 & 29.1 \\
Jiang \textit{et al.}~\cite{jiang2023probabilistic} & CVPR’23   & 24.0 & 25.4 & 26.6 & 30.4 & 32.1 & 20.1 & 20.5 & 36.5 & 40.1 & 29.5 & 27.4 & 27.6 & 20.8 & 24.1 & 22.0 & 27.8 \\
TesseTrack~\cite{reddy2021tessetrack}  & CVPR’21   &17.5 &19.6 &17.2 &18.3 &18.2 &17.7 &18.0 &18.0 &20.5 &20.3 &19.4 &17.2 &18.9 &19.0 &17.8 &18.7 \\
Wan\textit{et al.}~\cite{wan2023view} & CVIU’23   &19.5 &20.9 &19.5 &18.3 &21.1 &20.0 &17.9 &21.3 &23.9 &30.1 &21.6 &19.9 &18.9 &22.8 &19.5 &21.1 \\
\rowcolor{gray!20}
\hline
Ours &-   & \textbf{10.9} & \textbf{12.7} & \textbf{12.8} & \textbf{14.7} & \textbf{11.7} & \textbf{14.2} & \textbf{12.9} & \textbf{13.5} & \textbf{7.8} & \textbf{14.2} & \textbf{11.3} & \textbf{11.4} &20.3 &22.8 &20.5 &\textbf{13.7}\\
\hline

\end{tabular}}
\end{table*}

\begin{itemize}

    \item \textbf{Dual-view 3D Pose Estimation}~\cite{liu2021dual}\textbf{.} 
    This method proposes a camera-parameter-free dual-view 3D pose estimation model, which utilizes HR-Net for 2D keypoint detection and employs a 3D regression network. 
    Through a self-supervised training strategy, virtual views are generated based on orthogonal projections, allowing the model to learn spatial relationships and projection constraints.
    
    \item \textbf{Generalizable Human Pose Triangulation}~\cite{bartol2022generalizable}\textbf{.} 
    This method introduces a randomized triangulation framework to generate and score random 3D pose hypotheses. 
    It focuses on enhancing generalization across different camera configurations and datasets, demonstrating robust performance in multi-view pose estimation and related vision tasks.
    
    \item \textbf{LHPE-nets}~\cite{wang2022lhpe}\textbf{.} 
    This method adopts a weakly supervised framework to address the ambiguities in 3D pose estimation via multi-task learning. 
    It leverages data augmentation and latent constraints to enhance the robustness and generalization ability of the model across various datasets.
    
    \item \textbf{Probabilistic Triangulation for Uncalibrated Multi-view 3D Pose Estimation}~\cite{jiang2023probabilistic}\textbf{.} This method proposes a probabilistic triangulation module for uncalibrated multi-view 3D pose estimation. 
    It models the camera parameters using probability distributions and updates them through Monte Carlo sampling, thereby eliminating the need for calibration and improving the applicability in unstructured environments.
    
    \item \textbf{Self-supervised Multi-view Learning for 3D Pose Estimation}~\cite{jenni2020self}\textbf{.} 
    This method presents a self-supervised multi-view synchronization framework for learning 3D structure-aware features. 
    By detecting rigid transformations between image pairs, it utilizes large-scale unlabeled data for pre-training, thus improving the efficiency of labeled data usage.
    
    \item \textbf{Smart-VPoseNet}~\cite{wang2022smart}\textbf{.} 
    This method designs an intelligent viewpoint selection strategy called Smart-VPoseNet, combined with a viewpoint discriminative network. 
    By dynamically selecting high-quality viewpoints based on visibility, body stretch, and model affinity, it reduces errors caused by occlusion or poor camera angles.
    
    \item \textbf{View Consistency Triangulation for 3D Pose Estimation}~\cite{wan2023view}\textbf{.} 
    This method proposes a holistic triangulation framework that combines multi-view consistency and anatomical prior constraints. 
    By optimizing 2D keypoints through multi-view fusion and utilizing anatomical consistency, it reconstructs the complete 3D pose, enhancing the rationality and consistency of pose estimation.

\end{itemize}

Subsequently, we evaluate the model on the DA-3DPE dataset and test several existing baseline models. 
The training configurations for these models are set to their respective optimal training parameters. 
Below is a description of these methods:
\begin{itemize}
    \item \textbf{MTF-Transformer}~\cite{shuai2022adaptive}\textbf{.} This method proposes a Multi-view and Temporal Fusing Transformer (MTF-Transformer), which includes the Multi-view Fusing Transformer (MFT) and Temporal Fusing Transformer (TFT) modules. 
    It adapts to varying numbers of views and video lengths without camera calibration, focusing on fusing 2D pose features into 3D poses.
    \item \textbf{Cross View Fusion}~\cite{qiu2019cross}\textbf{.} 
    This method introduces a cross-view fusion mechanism and a Recursive Pictorial Structure Model (RPSM), which iteratively refines 3D pose accuracy by integrating multi-view 2D heatmaps. 
    It focuses on leveraging geometric relationships across views to achieve superior 2D and 3D pose estimation accuracy.
    \item \textbf{Probabilistic Triangulation}~\cite{jiang2023probabilistic}\textbf{.} 
    This method proposes a Probabilistic Triangulation module, which models camera pose using a probability distribution and employs Monte Carlo sampling to iteratively refine pose estimation. 
    It focuses on achieving robust 3D human pose estimation in uncalibrated scenes.
\end{itemize}

\begin{table}[t!]
\centering
\caption{THE 3D human pose estimation results evaluated on the Human3.6M dataset.}
\label{table2}
\resizebox{\columnwidth}{!}{%
\begin{tabular}{p{0.28\columnwidth}||p{0.15\columnwidth}| >{\centering\arraybackslash}p{0.1\columnwidth}| >{\centering\arraybackslash}p{0.14\columnwidth}}
\hline
\rowcolor{gray!20}
Method & Venue & MPJPE & P-MPJPE \\
\hline
\hline
Qiu~\textit{et al.}~\cite{qiu2019cross} & CVPR’19 & 31.2 & - \\
Ma~\textit{et al.}~\cite{ma2023self} & -  & 81.9 & 52.1 \\
Smart-VPoseNet~\cite{wang2022smart} & KBS’22 & 67.2 & 48.3 \\
Jenni~\textit{et al.}~\cite{jenni2020self} & ACCV’20 & 64.9 & 53.5 \\
TR~\cite{zhao2023triangulation} & NeurIPS’23 & 25.8 & - \\
VitPose~\cite{xia2022vitpose} & ICCC’22 & 17.0 & - \\
\rowcolor{gray!20}
\hline
\textbf{Ours} & - & \textbf{13.7} & \textbf{5.6} \\
\hline
\end{tabular}%
}
\end{table}

\subsubsection{Analysis of the Advantages of Multi-View Methods}
From the test results in Table~\ref{table1}, it can be seen that all multi-view methods outperform single-view methods. Multi-view methods can capture detailed body pose information from different directions, effectively overcoming the limitations of single-view methods in specific scenarios.
In single-view methods, due to the constraints of the view angle or occlusion, some key body parts are difficult to capture accurately, leading to larger pose estimation errors. 
For example, during the ``Pose'' action, a single view may only capture the front pose, while the body position and angle of the back may be occluded. 
However, using a multi-view model allows data to be obtained from multiple angles, such as the side and rear views, thereby compensating for the occluded or missing parts, providing a more complete and accurate 3D pose estimation.
Moreover, single-view methods are subject to unavoidable visual blind spots. 
For instance, during the "Walk" action, a single view may fail to accurately capture the height of the raised foot or the degree of knee bending, especially when the subject is walking sideways. In contrast, the multi-view model effectively reduces these blind spots by integrating data from different angles, enabling more precise capture of body movement details and producing more accurate pose evaluation results. 
The lower error in the ``Walk'' action exemplifies the advantage of multi-view methods.
Furthermore, single-view methods are more susceptible to environmental factors, leading to pose estimation bias. 
For example, in the ``Photo'' action, strong lighting may blur the edges of the body in the single view, affecting pose judgment. 
The multi-view model, however, can use data from other views to avoid the interference of strong lighting, thereby correcting such biases and improving the reliability of pose estimation. 
The fusion of multi-view data not only compensates for the blind spots of individual views but also allows for cross-validation of information, reducing errors caused by environmental factors or invalid data.
Finally, multi-view methods show significant advantages in capturing subtle pose variations. 
For example, in the ``Greet'' action, the small positional changes of the hands and head may be difficult to distinguish accurately in a single view. 
By combining multi-view data, the model can highlight these subtle changes and supplement feature representations from different angles, providing more precise results when evaluating complex actions.
In summary, multi-view methods overcome the limitations of single-view approaches by acquiring information from multiple angles, allowing for more accurate and comprehensive capture of body pose details, thereby improving the accuracy and reliability of 3D pose estimation.

\begin{table*}[!t]
\scriptsize
\renewcommand{\thetable}{III}
\centering
\caption{THE 3D human pose estimation results evaluated on the established DA-3DPE dataset.}
\label{table3}
\resizebox{\linewidth}{!}{
\setlength{\tabcolsep}{3pt}
\begin{tabular}{l||l|ccccccccccccccc|c}
\hline
\rowcolor{gray!20}
\textbf{MPJPE$\downarrow$}       &   \textbf{Venue}        & \textbf{Dir.} & \textbf{Disc.} & \textbf{Eat}  & \textbf{Greet} & \textbf{Phone} & \textbf{Photo} & \textbf{Pose} & \textbf{Purch.} & \textbf{Sit}  & \textbf{SitD.} & \textbf{Somke} & \textbf{Wait} & \textbf{WalkD.} & \textbf{Walk} & \textbf{WalkT.} & \textbf{Average} \\
\hline
\hline
Jiang \textit{et al.}~\cite{jiang2023probabilistic} & CVPR’23  &131.8 &135.9 &130.1 &122.5 &125.4 &129.7 &130.8 &131.3 &130.7 &131.1 &132.2 &129.7 &128.5 &131.1 &138.3 &130.4 \\
Shuaitextit {et al.}~\cite{shuai2022adaptive} &TPAMI’23    & 37.3 & 57.3 & 52.9 & 69.2 & 61.1 & 76.1 & 81.6 & 86.9 & 60.8 & 80.4 & 52.9 & 85.6 & 72.3 &93.2  & 113.5 & 72.1 \\
Qiu \textit{et al.} ~\cite{qiu2019cross} & CVPR’19     & 31.2 & 37.8 & 31.9 & 61.3 & 36.6 & 30.5 & 33.1 & 43.1 & 79.2 & 35.5 &46.2 & 58.5 & 29.3 & 38.5 & 30.3 & 41.4 \\
\rowcolor{gray!20}
\hline
\textbf{Ours} &-   &\textbf{13.6} & \textbf{9.5} & \textbf{14.5} & \textbf{14.0} & \textbf{12.8} & \textbf{19.1} & \textbf{14.7} & \textbf{11.1} & \textbf{13.0} & \textbf{27.6} & \textbf{12.8} & \textbf{9.2} & \textbf{16.5} & \textbf{29.3} & \textbf{29.9} & \textbf{15.9}\\
\hline
\end{tabular}}
\end{table*}

\subsubsection{Analysis of Evaluation Results on the Human3.6M Dataset}
Through comparison with other multi-view models, the testing results of DeProPose on the Human3.6M~\cite{ionescu2013human3} dataset indicate a significant improvement in accuracy across various actions compared to other models. 
The experimental test results in Table~\ref{table1} and Table ~\ref{table2} show that, whether for common actions (such as ``Phoning'', ``Eat'', ``Waiting'') or more complex actions (such as ``WalkDog'', ``Walking'', ``WalkTogether''), DeProPose effectively controls errors and achieves optimal performance.
For instance, in the ``Sit'' action, DeProPose achieves an error of only $7.8mm$, significantly lower than the errors of other comparative models. 
This demonstrates that DeProPose exhibits high precision in 3D human pose estimation for these actions. 
In the ``Dir'' action, DeProPose has an error of $10.9mm$, while other comparative models (such as Dual-view~\cite{liu2021dual}) have an error of $21.4mm$, and the Gener model has an error of $27.5mm$. 
This highlights DeProPose’s advantage in accurately estimating the 3D human pose for these actions.
Furthermore, from the perspective of stability and reliability, DeProPose maintains relatively consistent error rates across different actions. 
For example, in actions such as ``Phone'', ``Photo'', and ``Pose'', the errors remain between $11{\sim}13mm$, with no significant fluctuations. 
This indicates that the model is able to maintain consistent performance across different types of 3D human pose tasks, unaffected by variations in action types.
In summary, DeProPose provides accurate evaluation results when facing human poses in different scenarios, whether for static actions (such as ``Greet'' and ``Sit'') or dynamic actions (such as ``Walk'' and ``Wait''). 
This demonstrates the model’s strong robustness, enabling it to stably and accurately capture key 3D human pose information in complex and dynamic real-world scenarios, with strong adaptability to changes in action types or pose complexity.

\subsubsection{Analysis of Evaluation Results on the DA-3DPE dataset}

As shown in Table~\ref{table3}, our method demonstrates exceptional performance in the 3D human pose estimation task under deficiency-aware scenarios, exhibiting significant advantages over existing state-of-the-art methods, especially in complex interference conditions such as noise, occlusion, and missing viewpoints, where it still maintains the lowest MPJPE.
Specifically, our model achieves an average MPJPE of $15.9mm$, significantly lower than Shuaitextit~\textit{et al.}~\cite{shuai2022adaptive} ($72.1mm$), Qiu~\textit{et al.}~\cite{qiu2019cross} ($41.4mm$), and Jiang~\textit{et al.}~\cite{jiang2023probabilistic} ($130.4mm$), fully demonstrating the model's robustness and adaptability to disturbances.
This series of outstanding performances is attributed to the core innovation of our model—an adaptive feature fusion mechanism. Through dynamic weight adjustment based on projection and absolute errors, the model intelligently selects and integrates the most informative features in scenarios with incomplete or disturbed multi-view information, effectively mitigating the impact of interference and achieving stable, high-precision 3D pose prediction. 
Additionally, our method handles various types of interference (such as noise, occlusion, and missing viewpoints) within a unified framework, eliminating the need for separate training models for different scenarios.
This significantly reduces training time and resources, while ensuring consistency and efficiency of the model across different disturbance conditions.

Experimental results show that our unified model not only matches but even outperforms models trained independently for specific types of disturbances. Through the adaptive mechanism, which dynamically adjusts feature weights, the accuracy and robustness of multi-view feature fusion are significantly enhanced. 
This capability enables the model to exhibit high stability and consistency across noisy data, occluded data, and missing viewpoint data. It effectively weakens the impact of noise, avoids occlusions, and intelligently compensates for missing viewpoints, thereby maintaining exceptional performance in complex dynamic environments.

In summary, our method, with its innovative adaptive feature fusion mechanism and efficient unified framework, significantly improves the robustness, accuracy, and adaptability of 3D human pose estimation in deficiency-aware scenarios, providing an efficient and reliable solution for multi-view 3D pose recognition in complex environments.

\begin{table}[!t]
\centering
\caption{Ablation Study and Parameter Evaluation}
\label{ablation}
\resizebox{\columnwidth}{!}{%
\begin{tabular}{c||c|c|c||c|c}
\hline
\textbf{Module} & \textbf{Normal} & \textbf{No Fusion} & \textbf{Views} & \textbf{1} &\textbf{4} \\
\rowcolor{gray!20}
\hline
\hline
\textbf{MPJPE} & 15.9 & 31.9 & & 31.1 & 15.9 \\
\hline
\end{tabular}%
}
\end{table}
%\vspace{-0.5cm}

\subsection{Ablation Study}
As shown in Table~\ref{ablation}, the ablation study demonstrates the significant contribution of the fusion module to the model's performance. 
By comparing the MPJPE (Mean Per Joint Position Error) under the ``Normal'' setting and the ``No Fusion'' setting, it is evident that the fusion module notably enhances the prediction accuracy.
Specifically, the MPJPE is $15.9mm$ in the ``Normal'' setting, whereas it increases dramatically to $31.9mm$ in the ``No Fusion'' setting. 
The higher MPJPE indicates that the absence of the fusion module impairs the model's ability to effectively integrate multi-view information, leading to a significant degradation in the accuracy of human pose estimation.

These results underscore the critical role of the fusion module in aggregating and optimizing feature information. 
By adaptively integrating multi-view features, the module enables the model to comprehensively capture spatial information of human poses, thereby significantly improving prediction precision. Without this module, the model must rely on individual views or simple feature concatenation, which results in a notable drop in performance.

In summary, this ablation study highlights the essential role of the fusion module in multi-view 3D human pose estimation and validates its design's effectiveness and necessity.

\subsection{Parameter Analysis}
\begin{figure}[!t] 
    \centering
    \includegraphics[width=\linewidth]{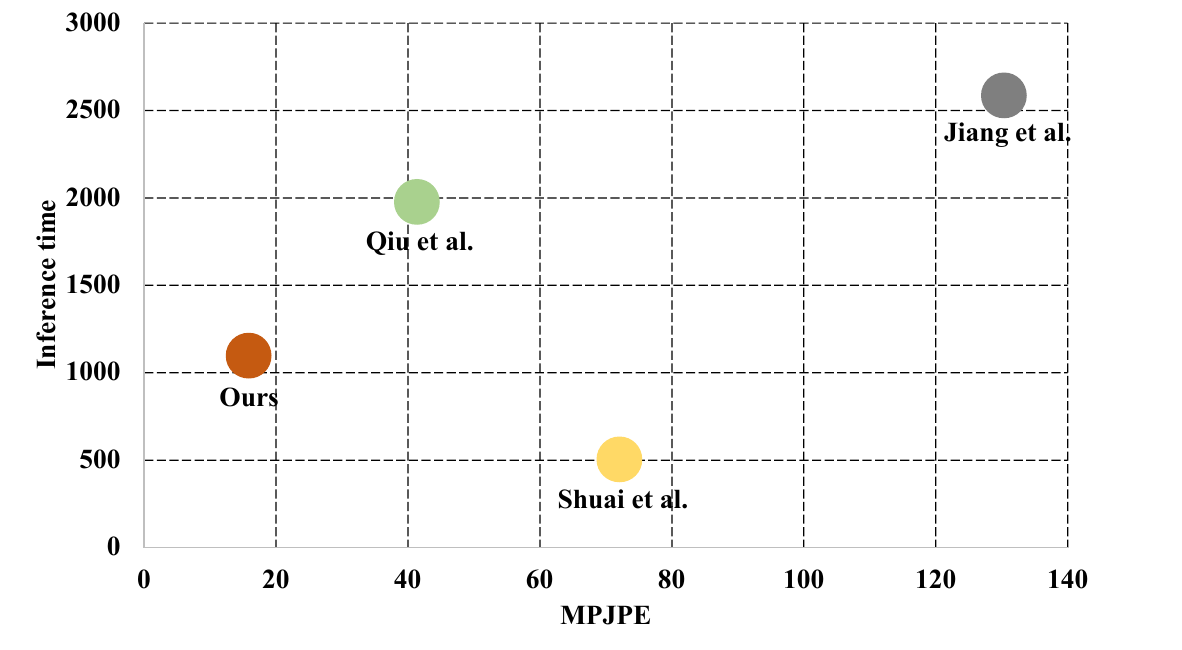} % 设置图片宽度
    \caption{Comparison of inference time \textit{vs.} MPJPE on S11 subject. Our model demonstrates significant efficiency improvements over the state-of-the-art methods, including Jiang~\textit{et al.}~\cite{jiang2023probabilistic}, Qiu~\textit{et al.}~\cite{qiu2019cross}, and Shuai~\textit{et al.}~\cite{shuai2022adaptive}, achieving lower inference time while maintaining competitive accuracy.}
    \label{时间图}
\end{figure}
\subsubsection{Time Comparison Analysis}
To evaluate the practical application potential of the model, we tested its inference speed. Specifically, we used all action images of the S11 subject as test data to assess the model's inference speed.
As shown in Fig.~\ref{时间图}, our model demonstrates significant efficiency advantages in terms of runtime. 
Compared to other methods, Jiang~\textit{et al.}~\cite{jiang2023probabilistic} required $2583.9s$, and  Qiu~\textit{et al.}~\cite{qiu2019cross} required $1975.1s$, while our model achieved a runtime of only $1095.7s$. 
This indicates that our model significantly improves runtime efficiency while maintaining high accuracy.
Although Shuai~\textit{et al.}~\cite{shuai2022adaptive} achieved a shorter runtime of $502.2s$, it is important to note that their method employs a two-stage process, and the reported time only includes the 3D inference stage, excluding the preceding 2D inference stage. 
Therefore, the overall efficiency advantage of our model becomes more evident in a complete end-to-end comparison. This efficiency can be attributed to several design optimizations, including the simplified network architecture, the computational efficiency of the Swin Transformer, and the optimization of the feature processing pipeline, which significantly reduces computational overhead. 
Such high efficiency not only underscores the theoretical significance of our model but also highlights its potential for real-time applications in practical scenarios.

\subsubsection{View Number Comparison Analysis}
The view number comparison experiment (Table~\ref{ablation}) further validates the importance of multiple views for model performance. 
When only a single view is used, the MPJPE is $31.1mm$, indicating a relatively large prediction error. 
In contrast, when four views are utilized, the MPJPE significantly decreases to $15.9mm$, demonstrating that incorporating multiple views substantially improves prediction accuracy. 
This improvement can be attributed to the diverse information provided by multiple views, which can compensate for the information loss caused by occlusion or noise in single-view settings. 
Additionally, our feature fusion module effectively integrates multi-view information, enabling more precise predictions of joint locations. 
In complex scenarios, the use of multiple views also enhances the model's robustness, ensuring stable high-accuracy predictions even under occlusions or missing data conditions. 
While incorporating additional views may introduce some computational overhead, our efficient fusion mechanism significantly mitigates this cost, achieving an optimal balance between accuracy and efficiency.

In summary, the parameter analysis experiments, through both runtime comparison and view number comparison, verify the superior performance of our model from different perspectives. 
On the one hand, the model demonstrates significant advantages in runtime efficiency, providing strong support for practical applications.
On the other hand, the model leverages multiple views to enhance prediction accuracy, showcasing exceptional adaptability and robustness, especially in complex scenarios. 
These experimental results highlight the innovative design of our model in theory and its high practical value in real-world applications.

\subsection{Visualization Results}
\begin{figure}[!t] 

    \centering
    \includegraphics[width=\linewidth]{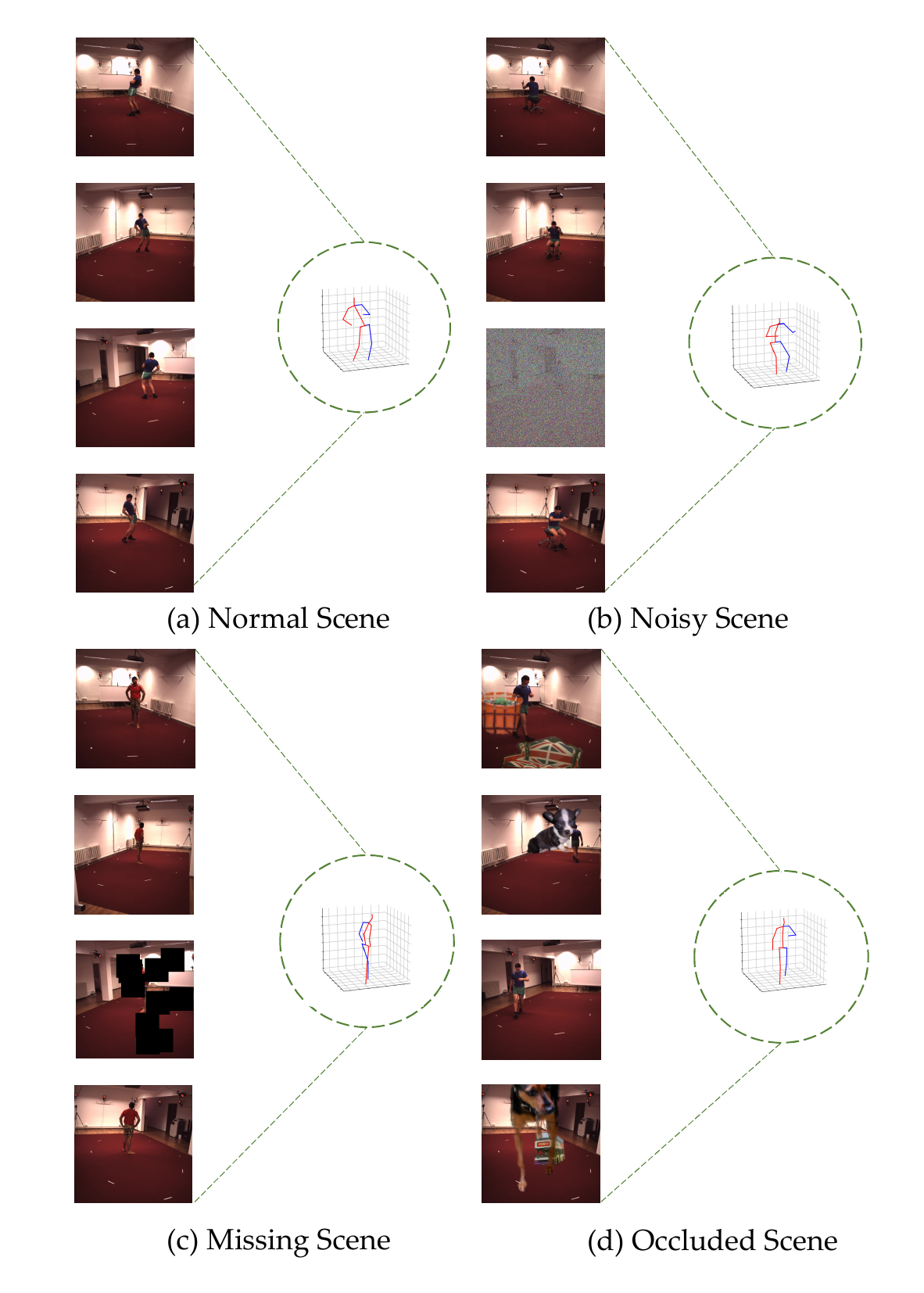} % 设置图片宽度
    \caption{The figure demonstrates the performance of our network in multi-view 3D human pose estimation across four scenarios: (a) normal scene, (b) noisy scene, (c) missing scene, and (d) occluded scene. 
    The results indicate that our network maintains high accuracy and robustness under various challenging conditions, highlighting its superior performance.}
    \label{可视化}
\end{figure}

As shown in Fig.~\ref{可视化}, we present the visualization results of multi-view 3D human pose estimation based on the DA-3DPE dataset. 
This dataset was specifically designed to address the challenges in multi-view 3D human pose estimation, with a focus on three key issues: missing data, noise interference, and viewpoint occlusion. 
These problems are commonly encountered in complex real-world scenarios and significantly impact the accuracy and robustness of existing methods. 
By comparing with normal images, we visually demonstrate the model's performance in dealing with these issues. The 3D human pose generated from four viewpoints clearly illustrates the model's adaptability and robustness when handling missing data, noise, and occlusion. 
Our visualization results not only highlight the unique advantages of the DA-3DPE dataset in simulating real-world problems but also showcase its enormous potential in multi-view 3D human pose estimation tasks. 
These results provide valuable insights for further optimization of the algorithm and demonstrate the efficiency and accuracy of our method in addressing real-world challenges.

\section{Conclusion}
The efficient end-to-end multi-view 3D human pose recognition model proposed in this paper demonstrates significant advantages in addressing challenges such as occlusion, noise interference, and viewpoint deficiencies in complex scenes. 
Unlike traditional methods that rely on multi-stage networks and module combinations, our model simplifies the network architecture, significantly reducing the difficulty of hyperparameter tuning while enhancing scalability. 
The core innovation lies in the development of a multi-view feature fusion mechanism based on projection and absolute errors. 
This mechanism adaptively assigns different weights to features from different views, accurately integrating information from multiple perspectives, thus effectively addressing occlusion and noise issues in multi-view complex scenarios. 
Furthermore, we generated a novel multi-view dataset that includes noisy and missing data, providing a foundation for comprehensive testing of the end-to-end multi-view 3D human pose recognition model. 
This dataset not only diversifies the testing scenarios but also advances research on occlusion issues in 3D human pose recognition. 
Experimental results show that despite the presence of various types of occlusions and noise in the dataset, the proposed model maintains high accuracy in complex scenes, demonstrating exceptional robustness and efficiency. 
This characteristic has broad application prospects in fields such as intelligent surveillance, motion capture, and virtual reality. 
By reducing the model's dependence on high-quality annotated data and optimizing the training process, our method not only achieves high efficiency in practical applications but also excels in handling occlusion, noise, and other challenges.

Future research can further extend the model to handle a broader range of occlusion types and explore ways to improve the model's accuracy and robustness in more complex poses and environmental variations. 
Moreover, with the continuous advancement of 3D pose recognition technology, further research will drive its widespread application in fields such as intelligent surveillance, virtual reality, and augmented reality.

\balance
\bibliographystyle{IEEEtran}
\bibliography{bib}

\end{document}